\documentclass[10pt,twocolumn,letterpaper]{article}
\usepackage[pagenumbers]{cvpr}

\usepackage{graphicx}
\usepackage{amsmath}
\usepackage{amssymb}
\usepackage{booktabs}
\usepackage{diagbox}
\usepackage{enumerate}
\usepackage{multirow}
\usepackage{lipsum}
\usepackage{cite}
\usepackage{graphicx}
\usepackage{amsmath}
\usepackage{amssymb}
\usepackage{booktabs}
\usepackage{diagbox}
\usepackage{enumerate}
\usepackage{multirow}
\usepackage{subcaption}
\usepackage{algorithm}
\usepackage{algorithmic}
\usepackage{times}
\usepackage{helvet}
\usepackage{courier}
\usepackage[hyphens]{url}
\usepackage{listings}
\usepackage{bbm}
\usepackage{xcolor}
\usepackage{indentfirst}
\usepackage{algorithm}
\usepackage{algorithmic}
\usepackage[pagebackref,breaklinks,colorlinks]{hyperref}
\usepackage[capitalize]{cleveref}
\crefname{section}{Sec.}{Secs.}
\Crefname{section}{Section}{Sections}
\Crefname{table}{Table}{Tables}
\crefname{table}{Tab.}{Tabs.}

\newcommand\blfootnote[1]{%
\begingroup
\renewcommand\thefootnote{}\footnote{#1}%
\addtocounter{footnote}{-1}%
\endgroup
}

\begin{document}

\title{TaCo: Textual Attribute Recognition via Contrastive Learning} 

\author{Chang Nie$^*$  \ \ \  Yiqing Hu$^{\dagger}$ \\
Yanqiu Qu  \ \ \ Hao Liu\ \ \  Deqiang Jiang\ \ \  Bo Ren\\
Tencent YouTu Lab\\
{\tt\small \{changnie, hooverhu, yanqiuqu, ivanhliu, dqiangjiang, timren\}$@$tencent.com}
}
\maketitle

\blfootnote{\hspace{-6mm} $^*$ Work completed during an internship at Tencent.\\
$^\dagger$ Corresponding author.\\
Preprint. Work in progress.
}

\begin{abstract}
As textual attributes like font are core design elements of document format and page style, automatic attributes recognition favor comprehensive practical applications.
Existing approaches already yield satisfactory performance in differentiating disparate attributes, but they still suffer in distinguishing similar attributes with only subtle difference.
Moreover, their performance drop severely in real-world scenarios where unexpected and obvious imaging distortions appear. 
In this paper, we aim to tackle these problems by proposing \textit{TaCo}, a contrastive framework for textual attribute recognition tailored toward the most common document scenes. 
Specifically, TaCo leverages contrastive learning to dispel the ambiguity trap arising from vague and open-ended attributes. 
To realize this goal, we design the learning paradigm from three perspectives: 1) generating attribute views, 2) extracting subtle but crucial details, and 3) exploiting valued view pairs for learning, to fully unlock the pre-training potential. 
Extensive experiments show that TaCo surpasses the supervised counterparts and advances the state-of-the-art remarkably on multiple attribute recognition tasks. 
Online services of TaCo will be made available.

\end{abstract}

\section{Introduction}
\label{intro}

\begin{figure}[t]
  \centering
   \includegraphics[width=1\linewidth]{ 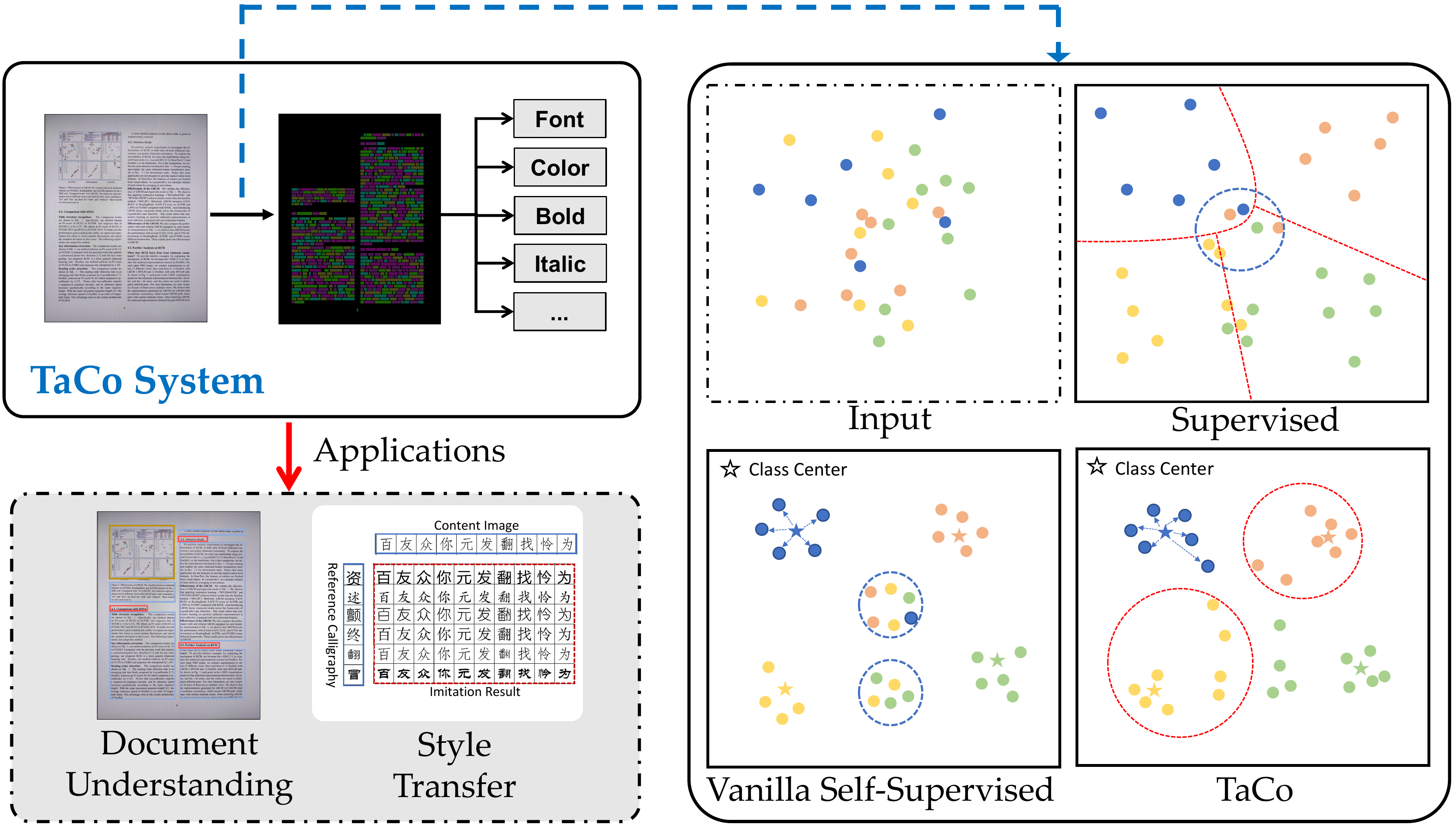}
   \caption{ (Left) Precious textual attributes benefit practical applications like document understanding and style transfer.
   (Right) Semantic spaces obtained from different learning paradigms. 
   Our TaCo system yields aligned attribute representations (red circle) for input with the same attributes beyond supervised approaches and vanilla self-supervised systems, which are constrained by label ambiguity (blue circle).}
   \label{img1}
\end{figure}

Textual attributes are fundamental in graphic design and also play a key role in forming document styles.
For example, in the case of converting a document image into editable formats like Microsoft Word~\cite{ref35}, retaining the original textual attributes is crucial for user experience. 
Moreover, graphic designers are keenly interested in identifying attractive styles, like word arts in the wild~\cite{ref14}. 
To achieve this goal, they may take photos of the target and turn to experts.
However, even for professionals, identifying the correct attributes from a combination of more than 1) 1000+ fonts~\cite{ref45}, 2) open-ended colors, and 3) other features is error-prone.
Hence, an accurate textual attribute recognition (TAR) system is expected to boost versatile applications, as shown in Fig.\ref{img1}.

\begin{figure*}[ht]
  \centering
   \includegraphics[width=1\linewidth]{ 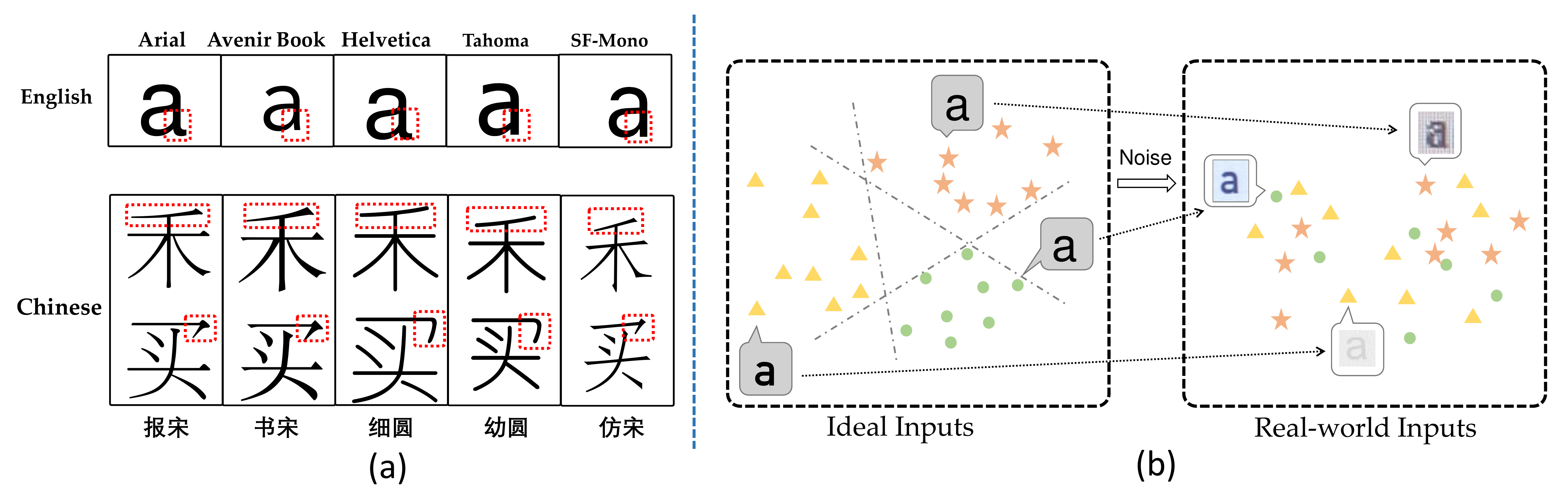}
   \caption{
   Challenges in textual attribute recognition.
   (a) Font attributes are rich and the distinction usually lies in local glyphs (red dashed boxes).
   (b) Existing approaches yield satisfactory result when inputted attributes are distinctive.
   However, real-world input introduces unexpected noises and makes it difficult for discrimination. 
   Best viewed in color.
   }
   \label{img2}
\end{figure*}

The design of TAR system is not a trivial task. 
The reason is mainly twofold: 1) \textit{\textbf{Unlimited attributes with subtle details.}}
Using the font attribute as an example, it is common to see that the basic difference between pairwise fonts reflected in subtle traits such as letter ending, weight, and slope~\cite{ref47}, as shown in Fig.~\ref{img2}(a).
As fonts are open-ended and ever-increasing through time, the continuously added new types intensified the recognition challenge~\cite{ref45}.
2) \textit{\textbf{Disparite attributes with similar appearance}}.
What is worse, the real-world input may not be ideal: even scanned PDFs and photographs may contain unexpected distortion that further blur the subtle traits.
As the consequence, the missing traits made the different attributes visually similar.
Existing methods~\cite{ref14,ref27} suffer in these complex scenarios, as shown in Fig.~\ref{img2}(b).
To mitigate these gaps, we propose TaCo, the first contrastive framework for textual attributes recognition.

\vspace{2mm}
\noindent\textbf{Technical Preview and Contributions.}
TaCo harnesses contrastive learning with elaborate pretext tasks to fulfill pre-training, allowing the model to learn comprehensive attribute representations over label-free samples.
The pretext tasks help to provide better attribute representation, especially for input with subtle changes and noises.
To further force the model to focus on local details, we introduce a masked attributes enhancement module (MAEM), which is achieved by dynamic feature masking operations together with a self-attentive mechanism.
MAEM also guides to learn from adjacent characters if existed, which built upon the fact that the attributes of adjacent characters are usually consistent. 
Finally, we introduce a paired-view scoring module (PSM) to guide the model to learn from high-quality attribute view pairs.
Using the generated comprehensive attribute representation as the backbone, we construct a recognition network to yield word-level and even character-level attribute recognition in complex real-world scenarios.
The contributions of this paper are summarized as follows:

\begin{itemize}
\item We propose a constrastive framework termed TaCo for textual attributes recognition.
TaCo is the first textual attribute recognition that supports multiple features at a time: 1) font, 2) color, 3) bold, 4) italic, 5) underline, and 6) strike.
The system could be easily extended to support incoming attributes.

\item By observing and leveraging the attribute-specific natures, we rigorously design the learning paradigm in respect of 1) generating attribute views, 2) extracting subtle but crucial details, and 3) automatically selecting valued view pairs for learning to ensure the effectiveness of pre-training.

\item Experimental results show the superiority of TaCo, which remarkably advances the state-of-the-art of multiple attributes recognition tasks. 
Online services of TaCo will be publicly released soon to assist relevant researchers and designers.
\end{itemize}

\begin{figure*}[ht]
  \centering
   \includegraphics[width=0.95\linewidth]{ 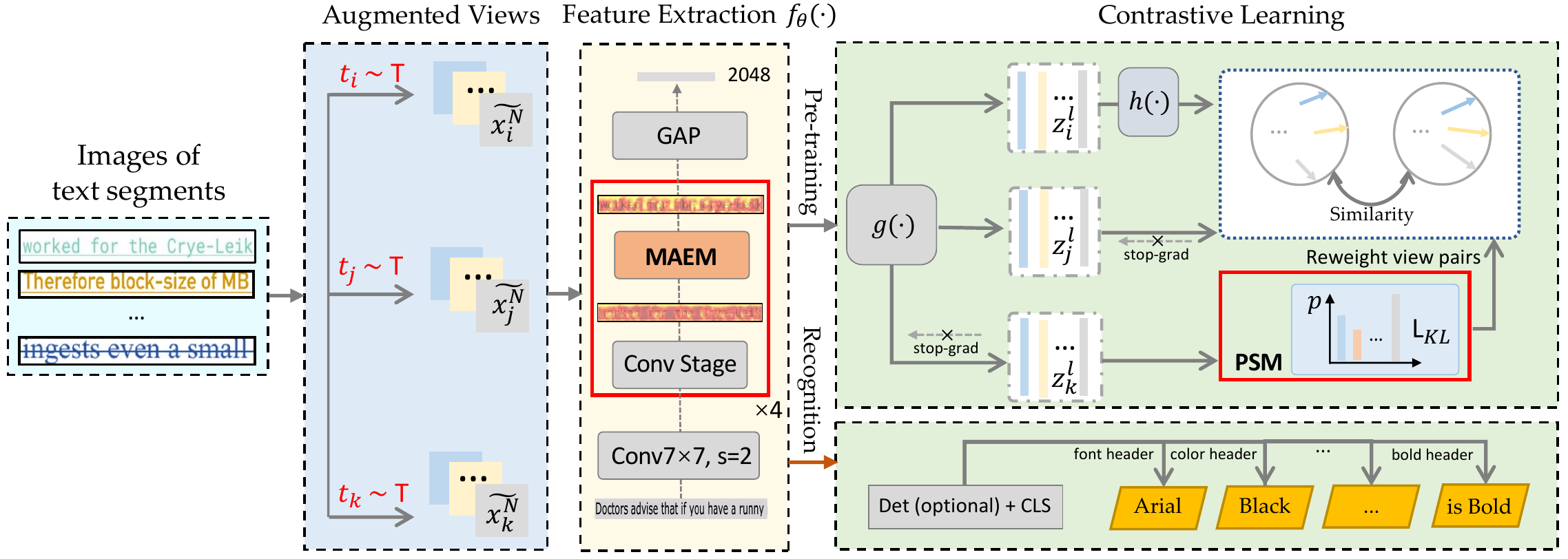}
   \caption{System overview of TaCo. 
   The pre-training pipeline of TaCo is built upon the SimSiam framework and consists of three key designs: 1) generation of augmented attribute views, 2) MAEM-guided feature enhancement, and 3) paired-view scoring module (PSM).
   $g(\cdot)$ and $h(\cdot)$ are an MLP head and a prediction MLP head, respectively. 
   For the recognition branch, an optional character detection module is introduced to provide character-level attribute recognition.
   Otherwise, TaCo outputs text-segments-level attributes by default.
   }
   \label{img3}
\end{figure*}

\section{Related Work}
\label{relwor}

\subsection{Textual Attribute Recognition}
Textual attribute recognition is essentially a fine-grained multi-tagging task, and plays a vital role in many scenarios. 
Unlike other typical text and entity categorization tasks, attributes rely less on language-specific semantics or layout knowledge, and more on the local details of words~\cite{ref14,ref40,ref47,ref64}. 
Several traditional methods~\cite{ref45,ref16} distinguish attribute classes heavily based on human-customized local feature descriptors and template annotation data, without generality and scalability for commercial applications. Recently, the upsurge of deep learning has dramatically advanced the development of TAR. DeepFont~\cite{ref14} firstly exploits CNN for font recognition and obtains favorable results. 
Moreover, Wang et al.~\cite{ref28} introduced transfer learning to address the domain mismatch problem between synthetic and real-world text images, as the prevalent of labeled attributes data scarcity. 
Chen et al.~\cite{ref27} designed a local enhancement module that automatically hides the most discriminative features during training to force the network to consider other subtle details. 
Unfortunately, existing supervised methods remain unsound in real-world scenarios since they failed to tackle label ambiguity and inter-class conflicts brought by image distortion.

\subsection{Self-Supervised Learning}
The self-supervised learning (SSL) allows the model to yield desirable representations from annotation-free data while relieving the burden of labeling~\cite{ref15,ref25}.
Consequently, pre-training based on joint multi-modal information has become the common practice for general-purpose document models. 
For example, the LayoutLM series~\cite{ref1,ref5,ref11} are pre-trained on the IIT-CDIP dataset containing 42 million images, which performs better than routinely training from scratch models. 
As is well known, the core of SSL involves designing proper pretext tasks and adopting the right evaluation criteria, \textit{e.g.}, masked signal recovery and visual token prediction in Visual-Language model~\cite{ref11}, and representation consistency of crafted views in contrastive methods~\cite{ref15,ref26}. 
Nevertheless, empirical evidence suggests~\cite{ref20} that existing models that learn from the whole documents are prone to capture global structured patterns without desired fine-grained stylistic features, thus inappropriate for attributes recognition tasks.

\section{Approach}
\label{Approach}
\subsection{Pre-training}

We adopt contrastive learning as the pre-training framework, which learns attribute information implicitly by comparing representations of pairwise positive views. 
As shown in Fig.~\ref{img3}, the TaCo system is built upon the SimSiam framework~\cite{ref26} and encapsulates three design refinements: 1) pretext task design, 2) masked attributes enhancement module (MAEM), and 3) paired-view scoring module (PSM).
The input of the pre-training system are images of text segments.
Text segments are a set of words with random length, which guarantee the sufficient context compared with a single character.
With this pre-training paradigm, TaCo is still able to recognizing attribute of a single character accurately, as shown in Section 3.2.

\vspace{2mm}
\noindent\textbf{Pretext Task Design.} 
The pretext task design (\textit{a.k.a.} data augmentation) is to construct suitable positive pairs of attributes.
For the attribute recognition tasks, they require neither semantic context nor relies on the global visual structure of the inputted images.
Hence, popular pre-training tasks including Masked Visual Language Modeling~\cite{ref5} and Gaussian blurring~\cite{ref15} are not suitable.
The former intends to learn from semantic while the latter affects the subtle and crucial feature of attributes.
We judiciously design the pretext tasks according to the nature of textual attributes.

Given an input image $x$, two separate operators $t_1, t_2$ randomly sampled from the augmentation family $\mathcal{T}$ are applied to $x$ to construct views $\widetilde{x}_i=t_i(x) $, $\widetilde{x}_j=t_j(x)$. 
For a \textit{minimal sufficient}\footnote{An optimal solution of $\mathop{\arg\min}_{f} I(f(\widetilde{x}_i); \widetilde{x}_i)$ is defined as the minimal sufficient encoder $f^*(\cdot)$ if $I(\widetilde{x}_i; \widetilde{x}_j)=I(f^*(\widetilde{x}_i); \widetilde{x}_j)$ holds~\cite{ref29}.} encoder $f(\cdot)$, the optimal $\mathcal{T}$ is supported to minimize
\begin{equation}\begin{split}
\mathbb{E}_{t_i, t_j\sim\mathcal{T}, x}\bigg[ & \underbrace{\Big |\Big|  I\big(f(\widetilde{x}_i); f(\widetilde{x}_j)\big) - I (\widetilde{x}_i; \widetilde{x}_j)\Big |\Big|}_{\#1} \\
& + \underbrace{d\big(f(x), f(\widetilde{x}_i)\big) + d\big(f(x), f(\widetilde{x}_j)\big)}_{\#2}\\
& + \underbrace{I(\widetilde{x}_i; \widetilde{x}_j) - \mathcal{H}(\widetilde{x}_i, \widetilde{x}_j)}_{\#3}
\bigg]
\label{eq1}
\end{split}\end{equation}
for $\forall\  t_i, t_j \in \mathcal{T}$ and $x$. 
Where $d(\cdot)$ denotes certain metrics, \textit{e.g.}, $\ell_1$ norm.
The first term $\#1$ in the expectation intends to reduce the noisy task-irrelevant mutual-information, and the remaining terms, $\#2$ and $\#3$, maximize the diversity of views with minimal task-relevant information. 
Hence, we empirically define pretext tasks consisting of three parts:
1) \textit{Random Cropping and Scaling} to take advantage of the content-dependent feature of attributes. 
The experiments reveal that making the views include varied textual content, or notably task-irrelevant information, is crucial for pre-training.
2) \textit{Color Jittering} is employed to prevent the network from learning trivial task solutions, such as color histogram features.
3) \textit{Random reordering} of characters to prevent the model from learning contextual semantic information.
This task is achieved by using synthetic training data introduced in Section 4.1.
The synthetic character-level bounding box enables this augmentation.

\vspace{2mm}
\noindent\textbf{Masked Attributes Enhancement Module.}
The Masked Attributes Enhancement Module (MAEM) is designed to achieve better attribute feature fusion in the encoder $f_\theta(\cdot)$.
The motivation of MAEM is that adjacent characters in one word share the same attributes with a higher probability.
Basing on this observation, MAEM incorporates dynamic masking operations and non-local attention mechanisms~\cite{ref38}, as shown in Fig.~\ref{img4}. 
Given a feature tensor $\mathcal{F}\in\mathbb{R}^{H\times W\times C}$, it is partitioned into non-overlapping patches and randomly masked with probability $p\sim\mathcal{U}(0, \delta)$, where $\delta$ is experimentally set as $0.2$. 
This step is performed towards the feature map rather than the input signal, and the masked position varies along the channel dimension. 
In contrast to \textit{dropout}~\cite{ref31}, we preserve the spatial continuity of unmasked patches, allowing the network to focus on details rather than global texture information. 
The mask operation is removed from the inference phase. 
Next, we utilize multiple convolutional blocks ($1\times1\  conv \rightarrow  BN \rightarrow  ReLU$) to avoid sharp edges caused by mask operations, and incorporate \textit{patchify} operation\footnote{The \textit{patchify} operation was initially designed to exploit the non-local self-correlation properties of infrared images~\cite{ref37}, and has recently been deployed in vision transformer and MLP architecture designs.} 
(\texttt{Split+Reshape}) to obtain the inputs, $ \mathcal{K},\mathcal{Q},\mathcal{V}\in\mathbb{R}^{h\times \frac{HW}{P^2}\times \frac{CP^2}{h}}$, of the self-attention mechanism. 
The final output feature tensor $\mathcal{G}\in\mathbb{R}^{H\times W\times C}$ can be computed by:
\begin{equation}\begin{split}
\mathcal{G} = {\left |  \right | }_{l=1}^{h} \Big( \hbar_P(\texttt{{Softmax}}( \frac{\mathcal{Q}_{l}\mathcal{K}_{l}^T}{\sqrt{CP^2/h}} )\mathcal{V}_{l})  \Big),
\label{eq2}
\end{split}\end{equation}
where $\left |  \right |$ denotes concatenation of $h$ attention heads, $\hbar_p$ means recovering feature maps from a sequence of patches of size $P\times P$. 
We embed the MAEM module into the encoder $f_\theta(\cdot)$ to make it more focused on local and contextual information.

\begin{figure}[t]
  \centering
   \includegraphics[width=1\linewidth]{ 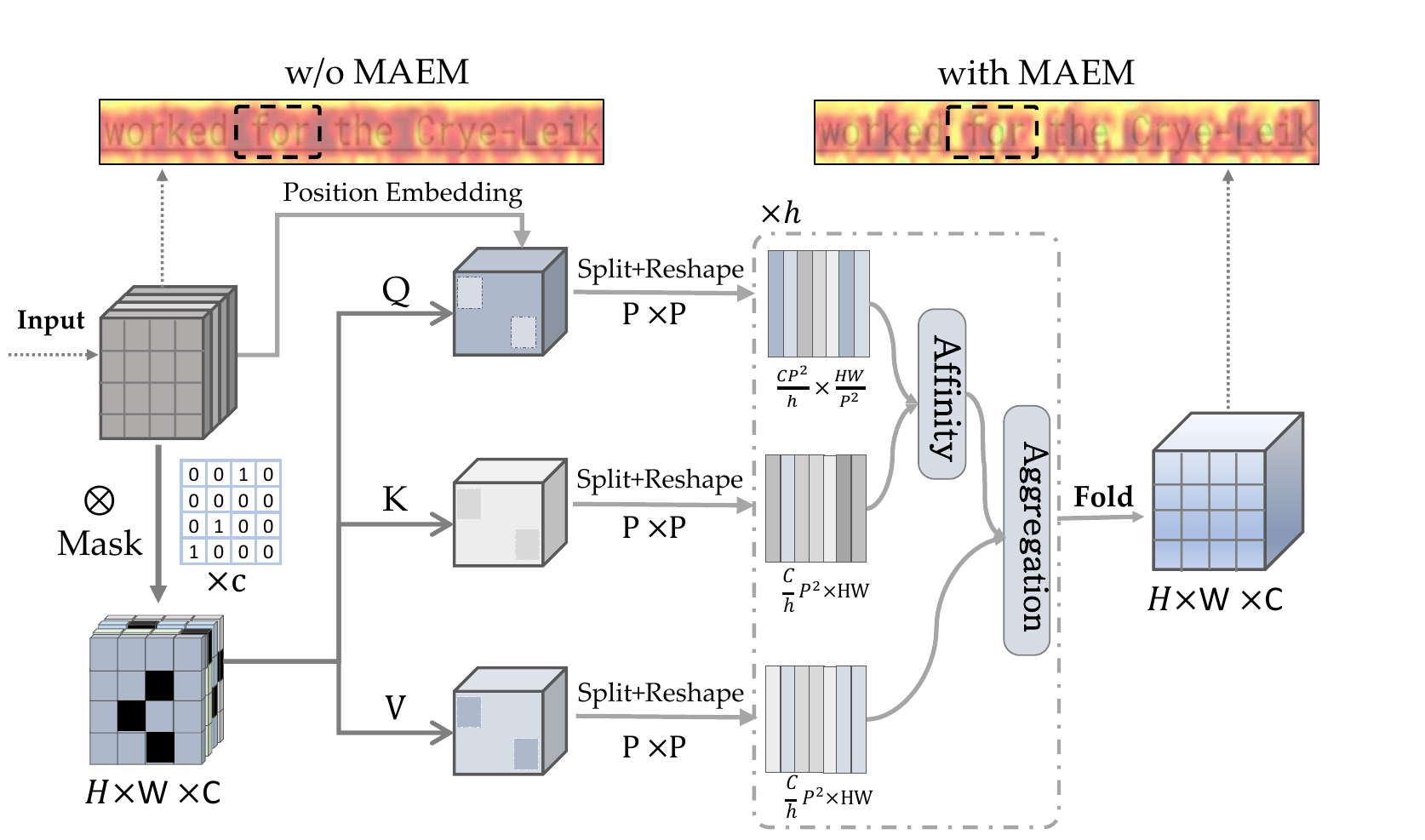}
   \caption{By feature masking and self-attention design, Masked Attributes Enhancement Module (MAEM) guides to learn local details for better attributes feature representation. The attention maps are acquired using Grad-CAM~\cite{ref63}. Best viewed in color and zoomed in.
   }
   \label{img4}
\end{figure}

\vspace{2mm}
\noindent\textbf{Paired-view Scoring Module.}
We design a Paired-view Scoring Module (PSM) to unleash the learning effectiveness of TaCo, which is parameter-free.
The motivation of PSM is that, owing to the randomness of sampling operations and inputs, the generated view pairs are not always guaranteed favorable, and low-quality views will impair the performance. 
For example, the view produced by random cropping may contain no words or only punctuation with incomplete attributes. 
In~\cite{ref30}, authors craft good positive pairs using the heat-map of a network to locate the regions of interest, but this relies on post-processing and handcrafted settings. 
For an original input image, PSM discriminates the quality of crafted pairs simultaneously during training.

Formally, given a batch of input images $\{x^l \}_{l=1}^N$, each sample is processed by three randomly sampled data operators $\{t_i^l, t_j^l,t_k^l \}\sim\mathcal{T}$ and the associated augmented views $\widetilde{x}^l_i,\widetilde{x}^l_j$, and $\widetilde{x}^l_k$ are obtained. 
Note that $t_k^l$ involves only color jittering to maintain view integrity. 
Then, a parametric function (\textit{e.g.}, ResNet-50) and a projection MLP head $g(\cdot)$ transform the views and feeds their representations $\{z_i^l, z_j^l, z_k^l\}$ into PSM. 
The computational flow can be presented as:
\begin{equation}\begin{split}
\textbf{Stage}\ \ \uppercase\expandafter{\romannumeral1}:\ \  \ &\widetilde{p_l}=\frac{1}{2} \big ( d(z_k^l, z_i^l) + d(z_k^l, h(z_j^l)) \big)\\
\textbf{Stage}\ \ \uppercase\expandafter{\romannumeral2}: \ & p_l=-|\widetilde{p_l}-\frac{1}{N}\sum_{l=1}^N \widetilde{p_l}|\\
& \mathcal{P}=\texttt{{Softmax}}(\{p_l/\tau)  \}_{l=1}^N),\\
\label{eq3}
\end{split}\end{equation}
where $\tau$ is a tuning temperature parameter, $h(\cdot)$ is a prediction MLP, and $d(\cdot, \cdot)$ denotes the cosine similarity defined as $d(x, y)=-\frac{x}{||x||_2}\cdot\frac{y }{||y||_2}$. 
On stage $\uppercase\expandafter{\romannumeral1}$, for each sample, we calculate the similarity of the intact view's feature $z_l^k$ with the two others. 
Clearly, when encoder $f_\theta(\cdot)$ is sufficient and the cropped views contain adequate or excessive task-relevant information, $\widetilde{p_l}$ up to a scale of 2. 
On stage $\uppercase\expandafter{\romannumeral2}$, We zero-meaned $\widetilde{p_l}$ within a batch and take the negative of its absolute value to measure the validity of each pair, where a smaller $p_l$ is better. 
In this way, the scoring mechanism forces the model to learn from pairs of moderate difficulty rather than those with excessively overlapping or incomplete content views. 
Then, a $\texttt{softmax}$ function is applied to normalize $p_l$ and output the pairs scores. 

In parallel, as shown in Fig.~\ref{img3}, the contrastive branch leverage a prediction MLP $h(\cdot)$ to transform the features of one view, and matches it to another one. 
The view-invariance of the system is reached by minimizing the cosine similarity of the pair representations, and a scored symmetric loss can be formulated as:
\begin{equation}\begin{split}
\mathcal{L}_{cos}=\frac{1}{2}\sum_{l} \mathcal{P}_l \big ( d(z_i^l, h(z_j^l)) + d(z_j^l, h(z_i^l)) \big),
\label{eq4}
\end{split}\end{equation}
where $\mathcal{P}_l$ is $i$th element of $\mathcal{P}$ in (\ref{eq4}).
Note that an important operation $stop$-$gradient$ is applied to $z_i,z_j$ before the gradient propagation. 
We introduce an additional Kullback-Leibler divergence loss to ensure the stability of pre-training, and the final optimization objective is derived as follows
\begin{equation}\begin{split}
\mathcal{L} = \mathcal{L}_{cos} + \lambda \frac{KL(\mathcal{R}|| \mathcal{P} )}{\log N},
\label{eq5}
\end{split}\end{equation}
where $\lambda$ is a trade-off constant, $\mathcal{R}$ denotes the expected uniform distribution of $\mathcal{P}$. 
The minimum possible value of $\mathcal{L}$ is $\lambda-1$. 
The whole framework is trained in an end-to-end manner. 

\subsection{TaCo: Attribute Recognition}
\label{sec:recog}

The final attribute recognition is built upon the backbone derived from pre-training.
Specifically, we train a multi-head linear classifier upon the backbone and apply it to recognize different attributes separately.
Now TaCo supports six textual attributes, namely 1) font, 2) color, 3) bold, 4) italic, 5) underline, and 6) strike.
More attribute types could be easily extended. 
The loss function is the sum of the cross entropy between the prediction and ground truth of all attributes, where the weight for font is set as $5$ and others as $1$, for font attribute learning is more difficult compared with other tasks.

Experimental results show that the TaCo system outperforms its counterparts by a large margin. 

As the default input of TaCo's pre-training and recognition is text-segments-level, 
it could be easily extended to character-level with a preposition character detection module. 
TaCo applies to complex cases with query images that present with totally different textual features. 
For example, for word ``TaCo'', character ``T'' maybe  ``$\mathbbm{T}$'' and character ``a'' maybe ``\textit{\textbf{a}}''. 
In this case, character-level attribute detection and classification is needed.
We leverage deformable DETR~\cite{ref34} as the detection framework, which outputs the bounding boxes and attributes for each character inside a query image. 
We take the last three stages outputs of the backbone as the input to the transformer head and fine-tune the whole network end-to-end until convergence.

\section{Experiments}
\noindent\textbf{Datasets.}
Now there exist no publicly available datasets for textual attributes. 
We constructed a large-scale synthetic dataset (SynAttr) comprising one million images of text segments for system pre-training and fine-tuning. 
One-tenth of the data contains words with more than two varying attributes for character-level attribute detection. 
For each sample, it contains words labeled with a bounding box and six attributes: font, color, italics, bold, underline, and strike.
For validation, we manually annotated a dataset Attr-$5$k comprising $5$k individual sentence images, which is cropped from 200 document images with various layouts and page styles collected from real-world scenes. 
More details of the datasets are given in the supplement.

\begin{table}[!t]
\centering
 \resizebox{.45\textwidth}{!}{
\begin{tabular}{ccc|c|c|c}
  \hline
  \textbf{Crop} &\textbf{Color} & \textbf{Shuffle} & \textbf{Pre. (\%) } & \textbf{Rec. (\%)} & \textbf{F1 (\%)}\\
  \hline
  &\checkmark&\checkmark&36.34 &33.05 &34.62 \\
  \checkmark&&\checkmark&53.40  &56.19   &54.76 \\
  \checkmark&\checkmark&&85.67  &90.22 &87.89 \\
  \hline
  \checkmark&\checkmark&\checkmark&\textbf{87.89}&\textbf{90.28}&\textbf{89.07}\\
  \hline
 \end{tabular}}
\label{tab1}
\caption{Linear evaluation under composition of data augmentations.  ``Shuffle'' refers to reordering the words, ``Crop'' means whether the generated views are a local region or full image, and “Color” represents color jittering. 
``Pre.'' and ``Rec.'' indicate average precision and recall.}
\end{table}

\begin{table}[!t]
\centering
 \resizebox{0.45\textwidth}{!}{
\begin{tabular}{l|c|c|c|c}
  \hline
  \textbf{Methods}   & \textbf{Pre. (\%) } & \textbf{Rec. (\%)} & \textbf{F1 (\%)} & \#{Params.}\\ 
  \hline
   ResNet-50 (vanilla)    &  85.35 & 84.21  & 84.78  & 23.60 M     \\
  \hline
    \ \ $\sim$ with SE ($r=16$) &85.93 {\color{gray} $_{+0.58}$}& 86.33 {\color{gray} $_{+2.12}$} & 86.13 {\color{gray} $_{+1.35}$}& 25.42 M\\
    \ \ $\sim$ with CBAM ($r=16$) & 86.62 {\color{gray} $_{+1.27}$} &86.94 {\color{gray} $_{+2.16}$} & 86.78 \textbf{{\color{gray} $_{+2.00}$}}& 25.63 M\\
  \hline
    \ \ $\sim$ with MAEM ($\delta=0$) & 86.69 {\color{gray} $_{+1.34}$} & 87.35 {\color{gray} $_{+3.14}$} & 87.02 {\color{gray} $_{+2.24}$}  & 23.81 M \\
    \ \ $\sim$ with MAEM ($\delta=0.2$) & \textbf{87.57} {\color{gray} $_{+2.22}$}& \textbf{87.96} {\color{gray} $_{+3.75}$} & \textbf{87.76} {\color{gray} $_{+2.98}$}  &  23.81 M\\
    \ \ $\sim$ with MAEM ($\delta=0.4$) & 86.88 {\color{gray} $_{+1.53}$} & 86.36 {\color{gray} $_{+2.15}$} & 86.62 {\color{gray} $_{+1.84}$}  &  23.81 M\\
    \ \ $\sim$ with MAEM ($\delta=0.6$) & 85.47 {\color{gray} $_{+1.12}$} & 86.46 {\color{gray} $_{+2.25}$}  & 85.96 {\color{gray} $_{+1.18}$} &  23.81 M\\
  \hline
 \end{tabular}}
\label{tab2}
\caption{Ablation study of the MAEM and comparison with two renowned attention modules. 
Each model is a single run from scratch.}
\end{table}

\begin{table}[!t]
\centering
 \resizebox{0.45\textwidth}{!}{
\begin{tabular}{l|c|c|c}
  \hline
  \textbf{Methods}   & \textbf{Pre. (\%) } & \textbf{Rec. (\%)} & \textbf{F1 (\%)}\\ 
  \hline
    Random init.            & 18.57  & 18.41  &18.49   \\
  \hline
    SimSiam (vanilla)       &87.89   &90.28   &89.07   \\
  \hline
    \ \ $\sim$ with Scoring ($\lambda=0$) & 87.18 {\color{gray} $_{-0.71}$} &  90.24 {\color{gray} $_{-0.04}$} & 88.68 {\color{gray} $_{-0.39}$}  \\
    \ \ $\sim$ with Scoring ($\lambda=0.2$) & 88.46 {\color{gray} $_{+0.57}$}& 91.28 {\color{gray} $_{+1.00}$} & 89.85 {\color{gray} $_{+0.78}$}\\
    \ \ $\sim$ with Scoring ($\lambda=2$) & \textbf{89.52} {\color{gray} $_{+1.63}$} & \textbf{91.34} {\color{gray} $_{+1.06}$} & \textbf{90.42} {\color{gray} $_{+1.35}$}    \\
    \ \ $\sim$ with Scoring ($\lambda=10$)   & 88.11 {\color{gray} $_{+0.22}$}  & 91.01 {\color{gray} $_{+0.73}$}  & 89.53 {\color{gray} $_{+0.46}$} \\

  \hline
 \end{tabular}}
\label{tab3}
\caption{Linear evaluation of the scoring module. 
``Random init.'' represents random initialization of model parameters instead of loading from pre-training.}
\end{table}

\begin{table*}[!t]
\centering
 \resizebox{0.98\textwidth}{!}{
\begin{tabular}{ll|ccccccc|ll}
  \hline
  &  Methods&  {Font} & Color & {Italic} & {Bold} & {Underline} & {Strike} & \textbf{Average}& \# {Params.} & \# {FLOPs} \\ 
  \hline
  \multirow{5}{*}{Baselines}& ResNet-50~\cite{ref32} & 86.71 & 95.78 & 98.72  & 90.40 & 87.10 &$\underline{99.15}$ & 92.98& 23.60 M&1.34 G \\
  &ResNeXt-101~\cite{ref44}&  87.33  & 94.10 &98.01 & 90.62 &86.95& 98.63 & 92.61 & 42.22 M& 2.62 G  \\
  &EfficientNet-b4~\cite{ref43}& $\underline{88.69}$ &\textbf{97.43} &$\underline{98.83}$ &91.00&$\underline{89.91}$&98.95& $\underline{94.12}$ &28.43 M&0.81 G \\
  &Swin-s~\cite{ref41} & 85.64 &$\underline{97.39}$ &97.51 &86.06 &84.73&97.62& 91.50 &48.75 M&8.51 G\\
  &CoAtNet-1~\cite{ref47} &87.39&96.15&97.96&87.84&84.31&98.72& 92.06 &33.05 M&6.81 G\\
  \hline
  \multirow{3}{*}{Variants} & DeepFont~\cite{ref14} & 88.07 & 96.09 & 98.28 &90.44 &86.26 &98.99 & 93.02 & 23.60 M& 1.34G \\
  & DropRegion~\cite{ref40}& ${88.42}$&96.29&98.18&$\underline{91.12}$&88.68&98.95& 93.61 & 37.88 M &6.45 G\\
  &HENet~\cite{ref27}& 87.90&95.68&98.48&89.67&88.30&99.03& 93.18 &23.60 M &1.38 G\\
  
  \hline
  \multirow{2}{*}{Ours} &TaCo w/o MAEM    & 93.25 & 97.22 & 99.01  &95.18&\textbf{90.51}& 99.13     & 95.72 & 23.60 M&1.34 G\\
  &TaCo    &  \textbf{94.28} &97.06  &\textbf{99.15} &\textbf{96.45} &89.45& \textbf{99.35} & \textbf{95.96} &23.81 M& 1.55 G\\
  \hline
 \end{tabular}}
\label{tab4}
\caption{Comparison with state-of-the-art recognition approaches. 
``w/o MAEM'' is short for “without using the MAEM". 
We achieved the best recognition performance over other baselines and variants.}
\end{table*}

\noindent\textbf{Implementation.}
The pre-training of our system is based on the SimSiam framework, with a backbone of vanilla ResNet-50   ~\cite{ref32}. 
The standard SGD optimizer with a learning rate of 0.1 is used for optimization. 
We train for 100 epochs (taking${\tiny \sim}$26 hours) and adjust the learning rate using a Cosine Annealing strategy. 
The patch size $P$ and the number of attention heads of MAEM are set to 4. 
For data augmentation, our pretext tasks include: 
1) randomly reordering the words with a probability of $0.5$, 
2) randomly cropping views from the original image by ratio range (0.8${\tiny \sim}$1, 0.6${\tiny \sim}1$, then rescaling and padding them to a fixed size of $(32, 256)$ without changing its aspect ratio, and 
3) color jittering alters the brightness, contrast, saturation and hue of an image with an offset degree of (0.4, 0.4, 0.4, 0.1) with a probability of $0.8$. 
In fine-tuning, we remove the data augmentations and retrain the whole network or multiple linear classifiers on frozen features until convergence. 
Moreover, we build a character detector upon backbone, which training follows the routine setup~\cite{ref34,ref20}. 
All experiments are implemented on a platform with 8 Nvidia V100 GPUs.

\begin{figure}[t]
  \centering
   \includegraphics[width=1\linewidth]{ 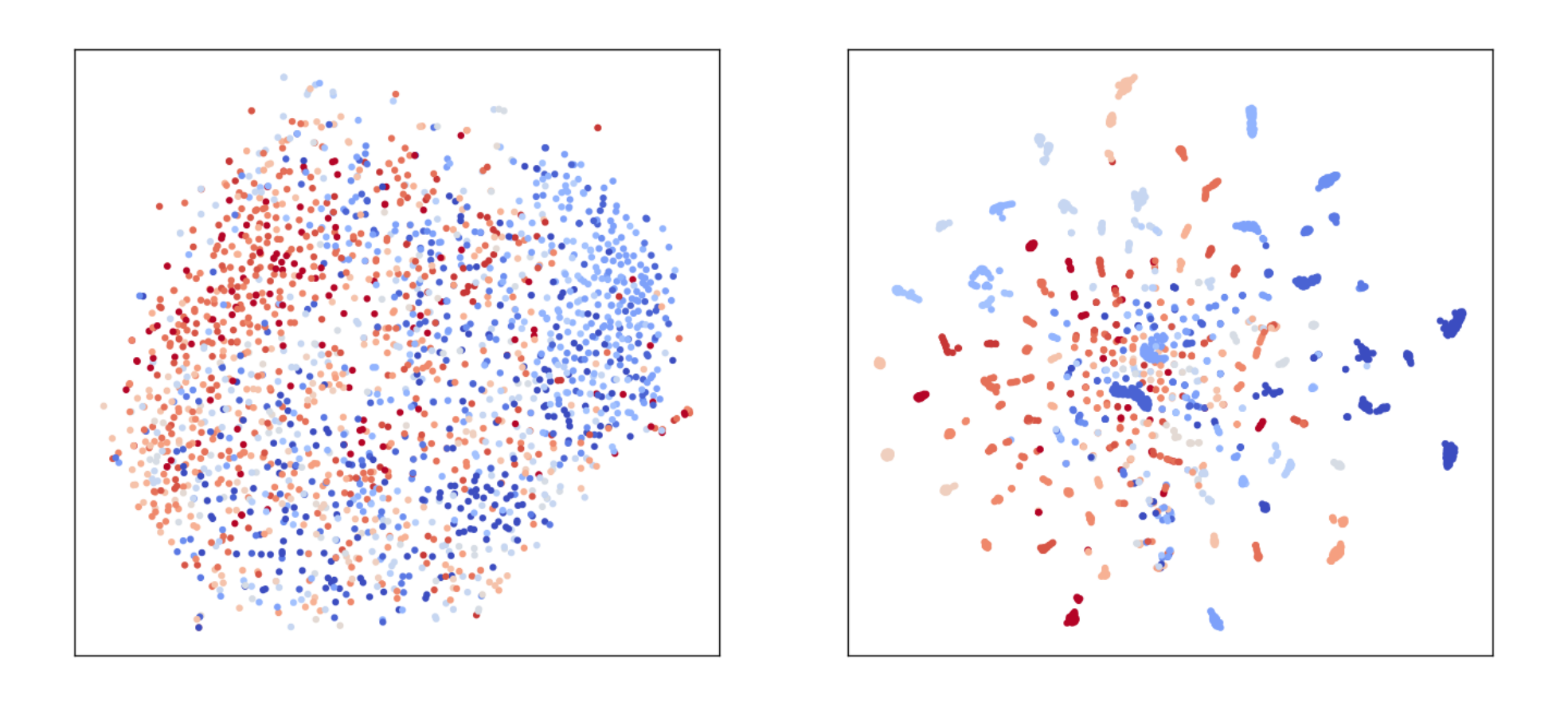}
   \caption{t-SNE visualization of features obtained by supervised learning (left) and TaCo (right). For visualization, we randomly selected 20 fonts from 2k random training images.}
   \label{img5}
\end{figure}

\subsection{Ablation Study and Analysis}
In this study, we conduct experiments on Attr-$5$k to investigate the contribution of individual components in our system. 
The font attribute is selected for ablation because its recognition is more difficult compared with others.
We take the precision, recall, and F1 score as the evaluation criteria.

\vspace{2mm}
\noindent\textbf{Pretext Tasks.}
We analyze the impact of each pretext task on performance. 
As shown in Table 1, we observe that removing ``Crop'' augmentation reduces precision, recall, and F1-score of font recognition by 51.55\%, 57.23\%, and 54.45\%, respectively. 
This concurs with the observation in~\cite{ref15}. 
As per predefined guidelines (Section 3.1), we argue that the ``Crop'' enables pair views to contain different content, thus reducing task-irrelevant mutual-information. 
For color jittering and words reordering, removal causes a decrease in the precision of 34.63\% and 2.22\%, respectively, which reflects the importance of view diversity. 
Overall, incorporating the three tasks yields favorable performance and guarantees the pre-training validity. 
The t-SNE results in Fig.~\ref{img5} show that the attribute representation shares better aggregation compared with supervised counterpart.

\vspace{2mm}
\noindent\textbf{Effect of MAEM.}
In the forward stage, the mask ratio of features at random is limited by a hyper-parameter $\delta$. 
In Table 2, the system equipped with MAEM brings a 2.98 F1-score gain when $\delta=0.2$, which introduces only 0.21M parameters. 
Furthermore, even if we set $\delta=0$, the non-local operation raises the recall by 3.14, showing the necessity to aggregate contextual information. 
We also compare MAEM with the reputed SE~\cite{ref60} and CBAM module~\cite{ref59}. 
Empirical results show that MAEM brings significant improvement with little memory overhead. 
We choose $\delta=0.2$ in the subsequent experiments.

\vspace{2mm}
\noindent\textbf{Effect of PSM.}
We conduct linear classification to show the benefit of the scoring module. 
As shown in Table 3, our system obtained varying magnitudes of improvements on the F1-score for different $\lambda>0$ settings. 
Notice that when $\lambda=0$, there is a slight drop in performance, which is probably attributed to the scoring trend shifting toward unsound samples. 
Notably, for $\lambda=2$, a precision gain of 1.63\% is obtained compared to the vanilla SimSiam, making a better trade-off between the scoring mechanism and the diversity of the learning samples.

\begin{table}[!t]
\centering
 \resizebox{0.45\textwidth}{!}{
\begin{tabular}{l|c|c}
  \hline
  \textbf{Methods} & \textbf{Pre. (\%) } & \textbf{Rec. (\%)} \\ 
  \hline
  Char-Cls~\cite{ref32} & 77.5 & 78.8  \\
    \ \ $\sim$ with pre-training & 89.2 & 88.6\\
  \hline
  Deformable DETR~\cite{ref34} &90.8&86.4 \\
    \ \ $\sim$ EfficientNet-b4~\cite{ref43} &91.4&89.8 \\
    \ \ $\sim$ RegNet~\cite{ref51} &92.1&90.4 \\
  \hline
  TaCo (Ours) &\textbf{96.9} & \textbf{93.6} \\
  \hline
 \end{tabular}}
\label{tab5}
\caption{Comparison with state-of-the-art methods on character-level font recognition. 
``Char-Cls'' refers to the categorization of each word region individually.}
\end{table}

\subsection{Comparison with the State-of-the-Art}
\noindent\textbf{Attributes Recognition.}
We experimentally demonstrate that the pre-training yields remarkable performance gains for attributes recognition, especially for font. We fine-tune the pre-trained encoder and output the recognition results for six attributes with multiple classifiers. 
Several strong baselines and other modified variants are choosing for comparison.
The evaluation metric is average recognition accuracy. 
For fairness, all models are trained on the same datasets and settings. 
Table 4 presents the evaluation results of all methods on Attr-$5$k. 
Specifically, the pre-trained ResNet-50 performs far beyond its peers, including EfficientNet~\cite{ref43} and variant HENet~\cite{ref27}, with 5.59\% and 5.86\% advantages in recognition performance for font. 
Besides, slight improvements are achieved for other attributes, such as italic and strike. 
We notice that deeper models (ResNeXt-101) and vision transformers (Swin-S) do not have obvious gain.
It is perhaps owing to attribute recognition is focusing more on local details rather than semantic interactions of different regions. 
As color jittering is crucial for other attributes, the slight inferior performance of color recognition is acceptable for TaCo. Detailed discussion is listed in the supplement. 

Overall, our average recognition accuracy over all attributes improves by 2.98 relative to the vanilla ResNet-50 and outperforms its counterparts significantly. 
This suggests the superiority of our pre-training regime. 
Albeit our system is learned in terms of fixed categories, the buildup pre-training pipeline allows it to scale to the newly designed classes.

\begin{figure}[!t]
  \centering
   \includegraphics[width=1\linewidth]{ 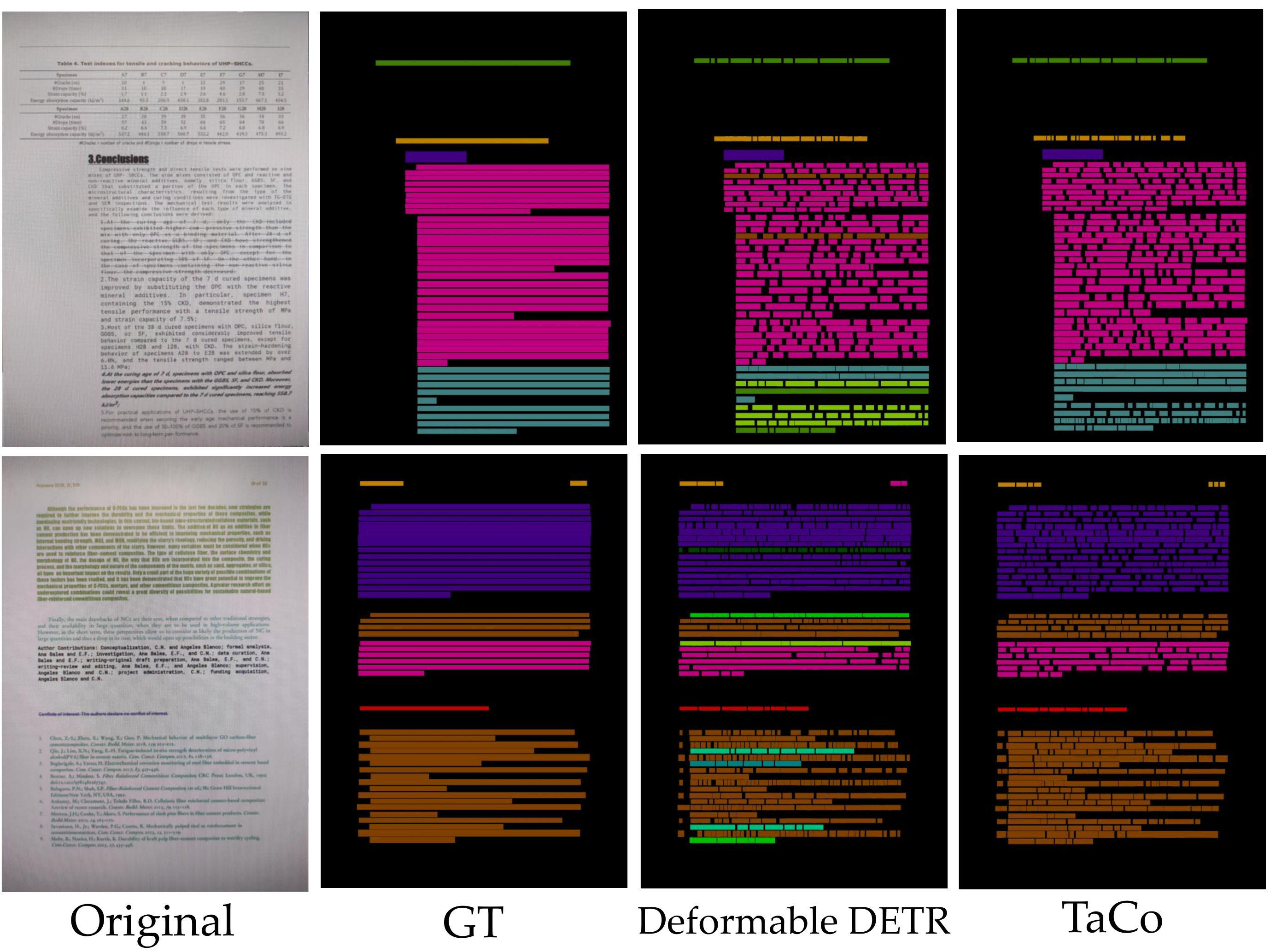}
   \caption{Visualization of character-level font recognition in ordinary document scenarios.}
   \label{img6}
\end{figure}

\noindent\textbf{Character-level Attributes Detection.}
We trained the optional character detector to support character-level attribute recognition.
The same training data and settings are reused. 
Table 5 shows the average font recognition precision and recall towards various approaches, where detection-based methods are measured with IoU = 0.5. 
Note that ``Char-Cls'' means that words are recognized sequentially based on available bounding boxes, and the input is a localized single-word region instead of a whole image.
We used deformable DETR as the benchmark and verified the behavior of different backbones separately. 
We observe that character-wise classification yields poor performance, 1.6\% precision lower than the deformable DETR, even with pre-training loaded. 
This suggests that contextual information benefits the words with indistinct features in the lexicon. 
For single-stage approaches, our system delivers an accuracy improvement of 5.1\% when loading backbone weights, which is more effective than replacing with a stronger baseline. 
Fig.~\ref{img6} visualizes the font recognition results for two real-world document images. 
The TaCo system achieves better accuracy than the supervised counterparts.

\begin{figure}[!t]
  \centering
   \includegraphics[width=1\linewidth]{ 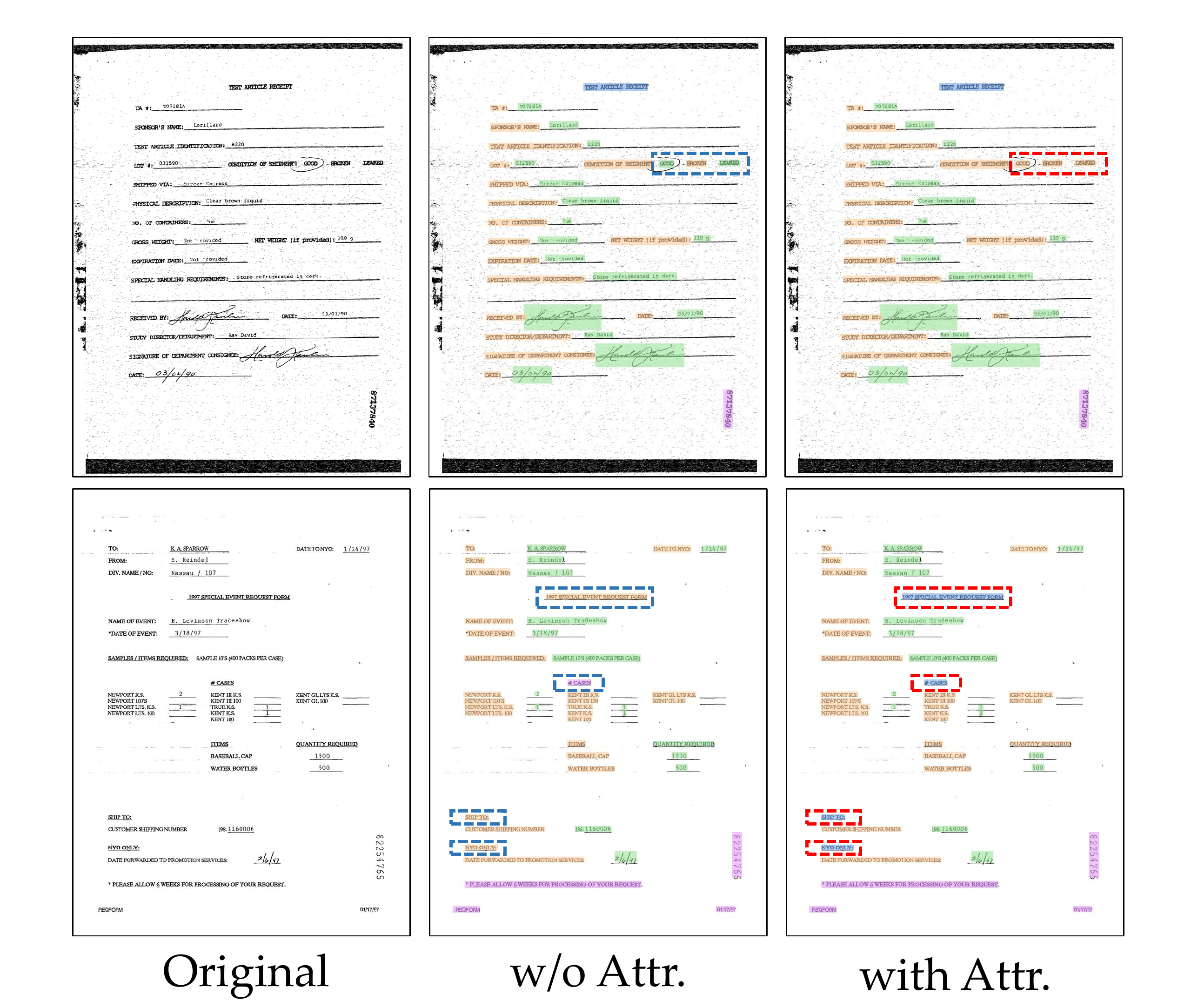}
   \caption{Visualization of document entity recognition. 
   Entities of the same classes are correctly identified with the aid of attribute information (red dashed boxes). 
   ``w/o Attr.'' is short for ``without using textual attribute modality''.
   }
   \label{img7}
\end{figure}

\subsection{Broader impact of TaCo}

\vspace{3mm}
\noindent\textbf{Semantic-Entity Labeling in Document.}
We validate the benefits of the additional provided textual attributes by TaCo on the Form Understanding in Noisy Scanned Documents (FUNSD) dataset~\cite{ref7}, which is a well-known challenging task in document understanding.
Specifically, we use TaCo to retrieve the attribute information of the text inside an image and embed it into a 512-dim linear space. 
Then, we construct a 2D attributes grid and sum it with stage-2 output features of ResNet-50. 
As shown in Fig.\ref{img7}, we can correctly identify entities with the same attributes, such as bolder header and underlined answers, and improve the precision by 1.49. 
More details are given in the supplement.

\begin{figure}[!t]
  \centering
   \includegraphics[width=1\linewidth]{ 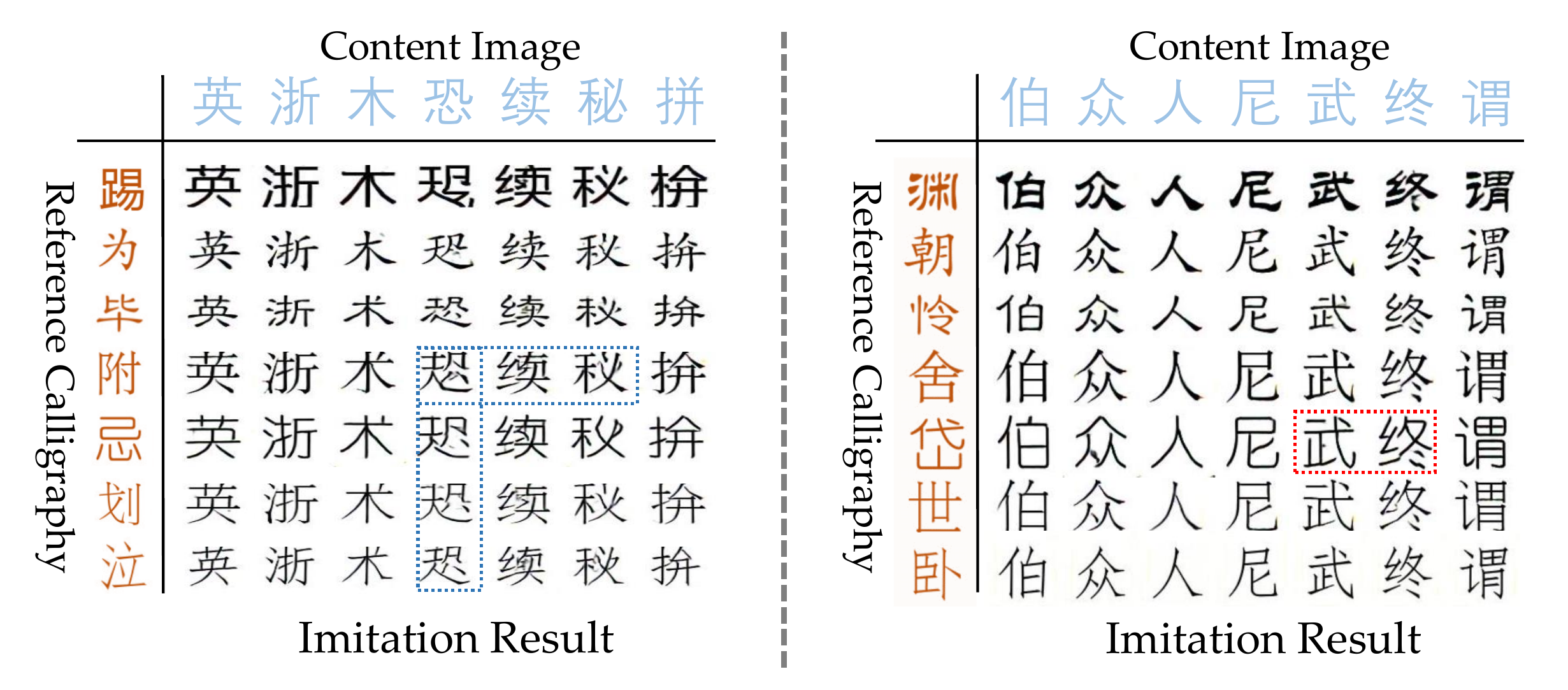}
   \caption{Visualization of font generation. The discriminator of DG-Font with loaded pre-training weights (Right) improves the fidelity of the generated font samples.}
   \label{img8}
\end{figure}

\noindent\textbf{Font Generation} aims to transfer the style of a reference calligraphy image to ones with different style~\cite{ref62}, thereby producing characters of a specific font. 
Existing Method like DG-Font~\cite{ref61} already yield satisfactory result, and their performance could be further boosted by treating TaCo as a friend.
We observe that these adversarial generative approaches require a powerful discriminator for identifying the font attributes of the synthetic and real samples. 
Hence, we take the pre-trained encoder of TaCo as the discriminator of DG-Font.
As shown in Fig.~\ref{img8}, the generated results of the model with loaded features are more realistic.

\section{Conclusion}
This paper presents a novel contrastive framework TaCo for retrieving multiple textual attribute information. 
By incorporating the attribute-specific characteristics, we rigorously design a pre-training pipeline based on contrastive learning with customized designs to warrant learning effectiveness. 
Experimental results suggest that our TaCo system is able to learn subtle but crucial features and exhibits superior performance against strong baselines.
For future research, we plan to support richer attributes like language classes.

\section*{Appendix}
\subsection*{A. Datasets Description}

\noindent\textbf{Synthetic Text Segments.}
All synthetic images of text segments are generated by writing plain texts collected from website into an editable structured document with predefined attributes. 
Fig.1 presents eight samples, where four complex cases (bottom) contain words with more than two varying attributes for character-level detector training. 
Each word inside an image is labeled with a bounding box and six attributes: font, color, italics, bold, underline, and strike.
The font covers 26 standard Chinese and 41 English categories, and their corresponding font libraries are available at FounderType\footnote{https://www.foundertype.com/}.
The color attribute covers 14 classes, and four others are labeled with True or False.

\begin{figure}[!h]
  \centering
   \includegraphics[width=0.9\linewidth]{ 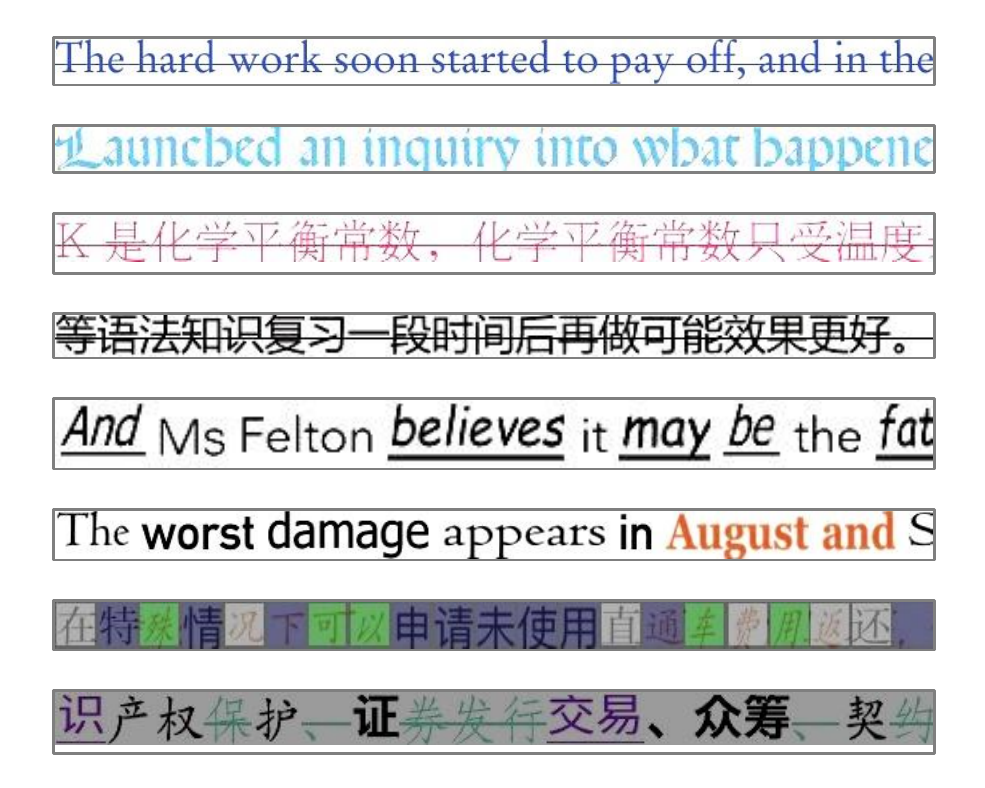}
   \caption{Synthetic samples of text segments.}
   \label{img8}
\end{figure}

\vspace{2mm}
\noindent\textbf{Attr-5k dataset.}
As mentioned in the main text, now there exist no publicly available datasets for textual attributes.
Therefore, we introduce a new large-scale benchmark called Attr-5k, which comprises 5000 images of text segments gathered from real-world document scenes. 
For diversity, the dataset involves various layouts and styles in English and Chinese, like advertisements, papers, reports and other wild structured scenarios.
Fig.3 (top row) shows four collected documents. Two types of annotations are included for each image: 1) Contents and bounding boxes of text segments. 2) Textual attribute labels, used to validate the attribute recognition approaches.

\subsection*{B. On the Performance of Color}

One may argue that color jittering limits the system from learning low-level color features during pre-training, thereby hampering the performance of color recognition. 
Notice that the pre-training pipeline of TaCo is designed to allow the system to better characterize local details such as letter weight and slope. 
This capability, in turn, compensates for color recognition in fine-tuning and brings a 1.44\% improvement in accuracy (Table 4 in the main text). 
Hence, we believe that color jittering is crucial for the pre-training.

\begin{figure}[ht]
  \centering
   \includegraphics[width=1\linewidth]{ 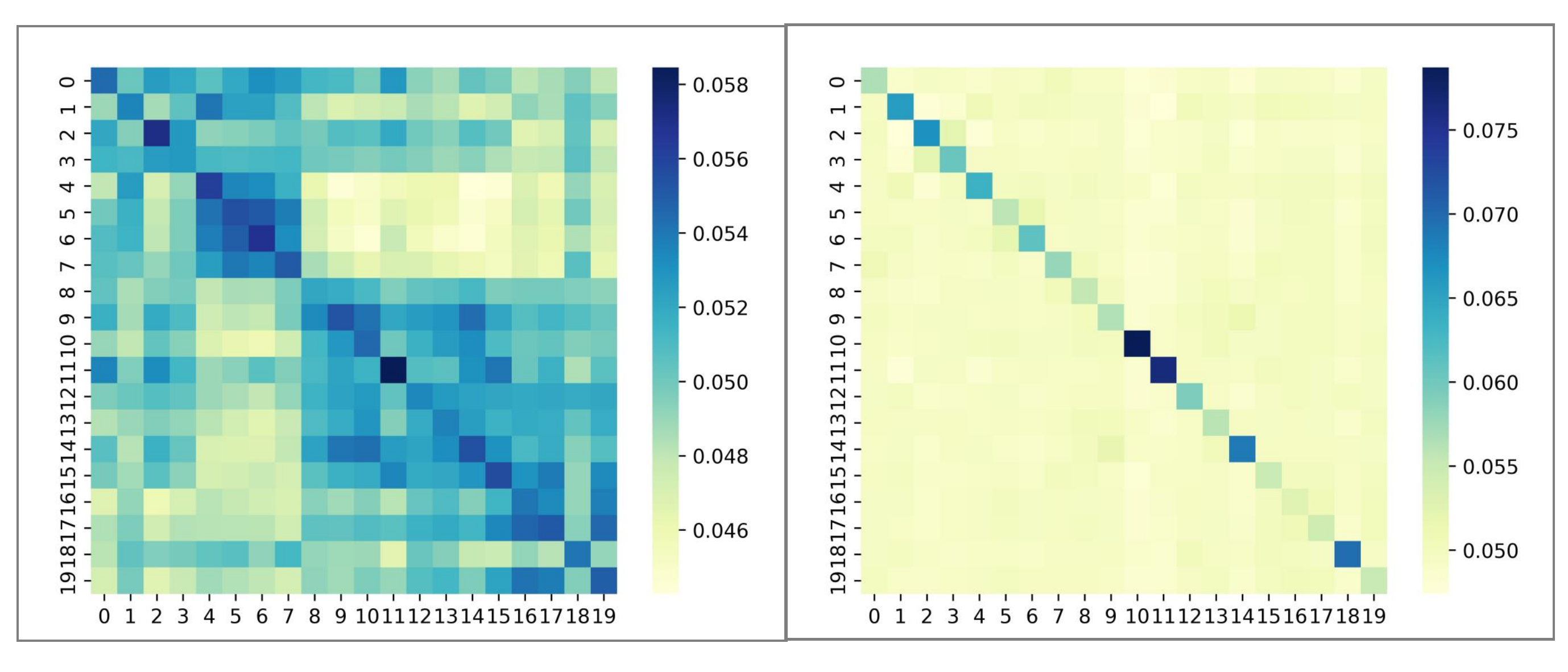}
   \caption{Difference of correlation matrices of features obtained by supervised learning (left) and TaCo (right). Obviously, TaCo leads to smaller inter-class correlations.}
   \label{img3}
\end{figure}

\begin{figure*}[ht]
  \centering
   \includegraphics[width=1\linewidth]{ 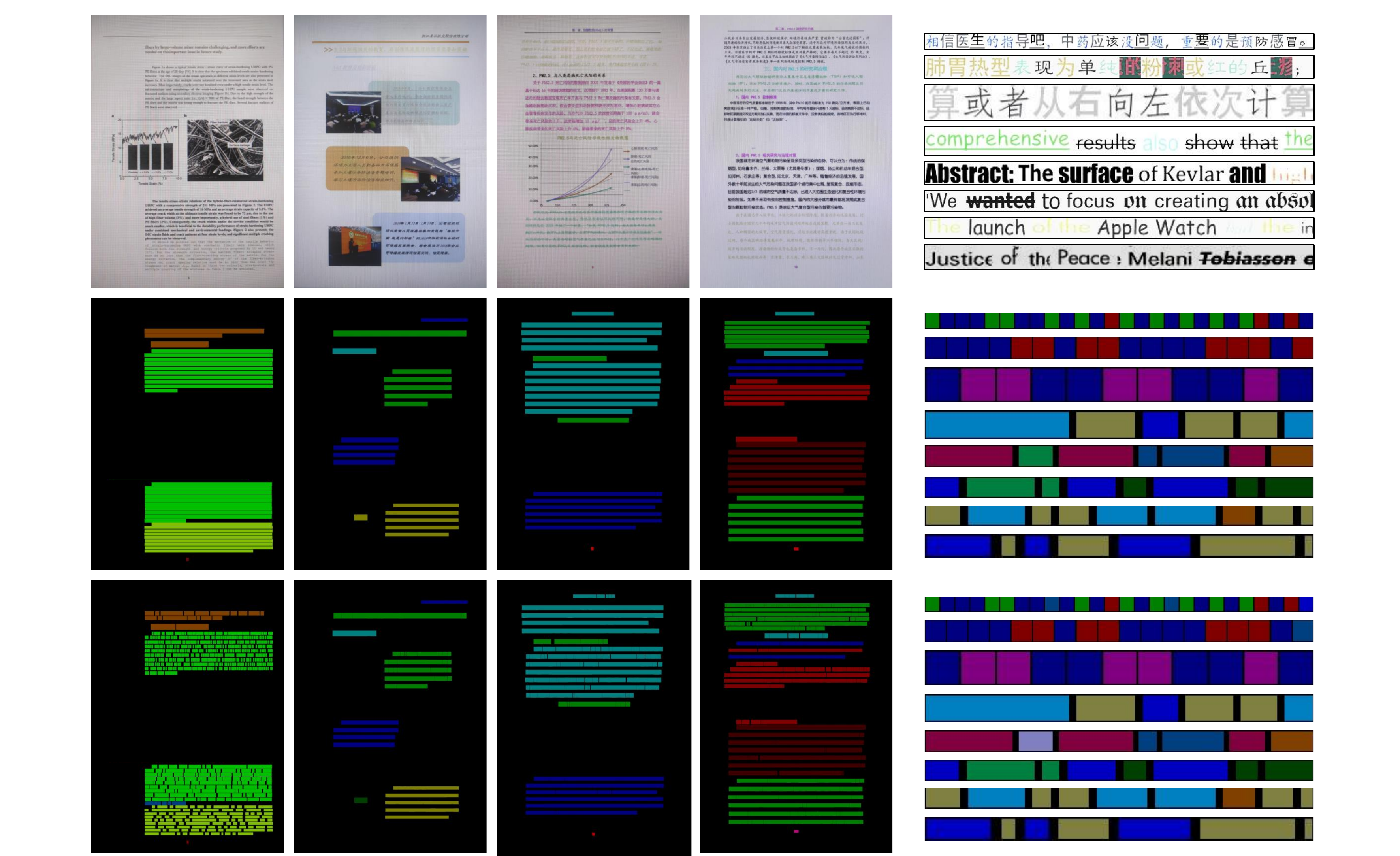}
   \caption{Visualization of character-level recognition results in ordinary and complex scenarios. For the given query images (top row), the TaCo system returns the character-level suggestions (bottom row, font only). The middle row is the ground truths.
   }
   \label{img3}
\end{figure*}

\begin{figure*}[ht]
  \centering
   \includegraphics[width=.9\linewidth]{ 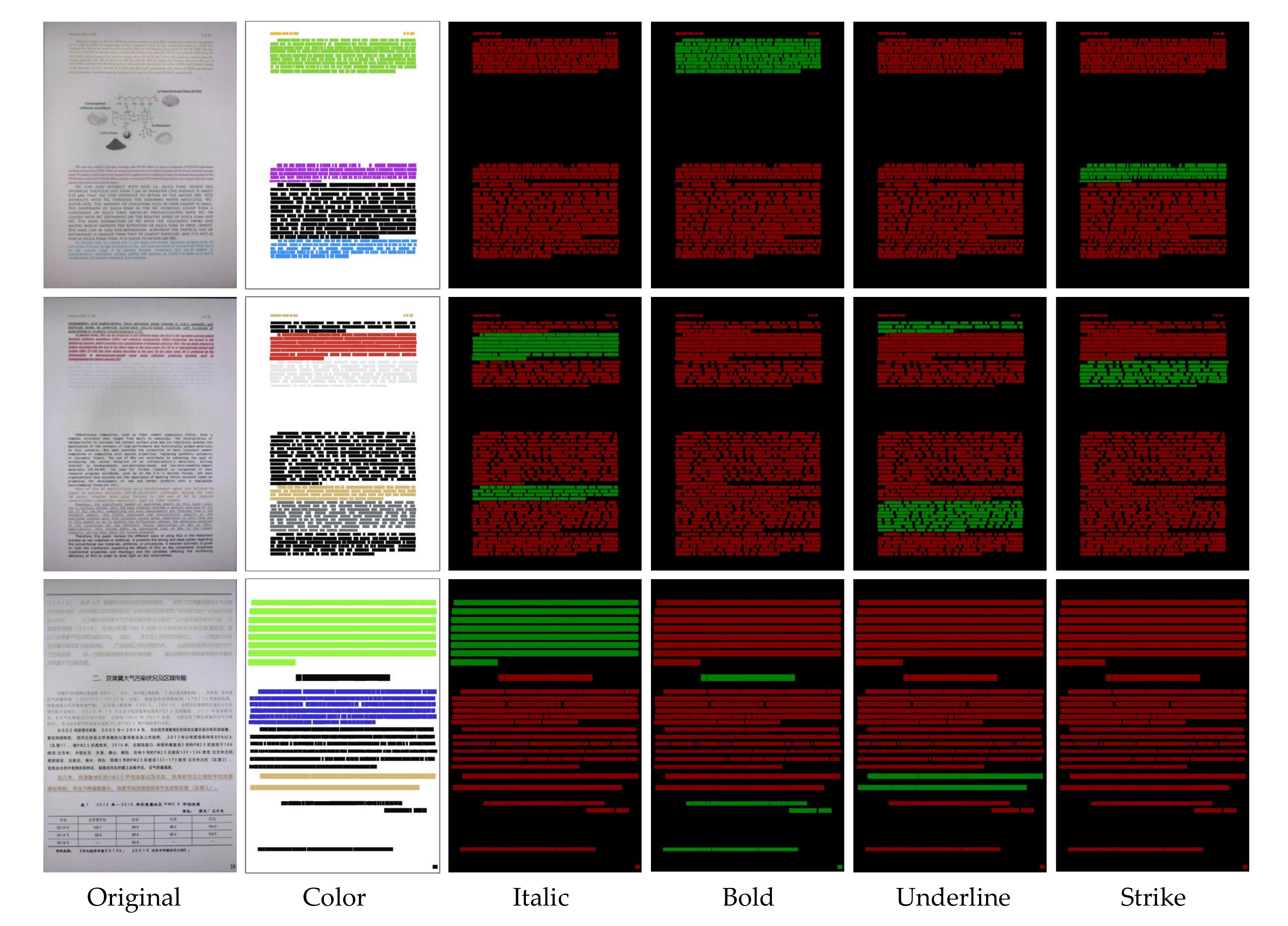}
   \caption{Visualization of character-level recognition results for five attributes (except font) in ordinary document scenarios. For the last four columns, green denotes true and red denotes false.}
   \label{img4}
\end{figure*}

\begin{table}[!h]
\centering
 \resizebox{0.45\textwidth}{!}{
\begin{tabular}{l|c|c}
  \hline
  \textbf{Methods} & \textbf{Pre. (\%) } & \textbf{Rec. (\%)} \\ 
  \hline
  ResNet-50~\cite{ref32} & 82.42 & 80.45  \\
  ResNet-50 with Attr. & 83.91& 82.60  \\
  \hline
 \end{tabular}}
\label{tab5}
\caption{Performance of Semantic-Entity Labeling on FUNSD, where Faster-RCNN~\cite{ref48} is used with different backbones.}
\end{table}

\begin{figure*}[!h]
  \centering
   \includegraphics[width=1\linewidth]{ 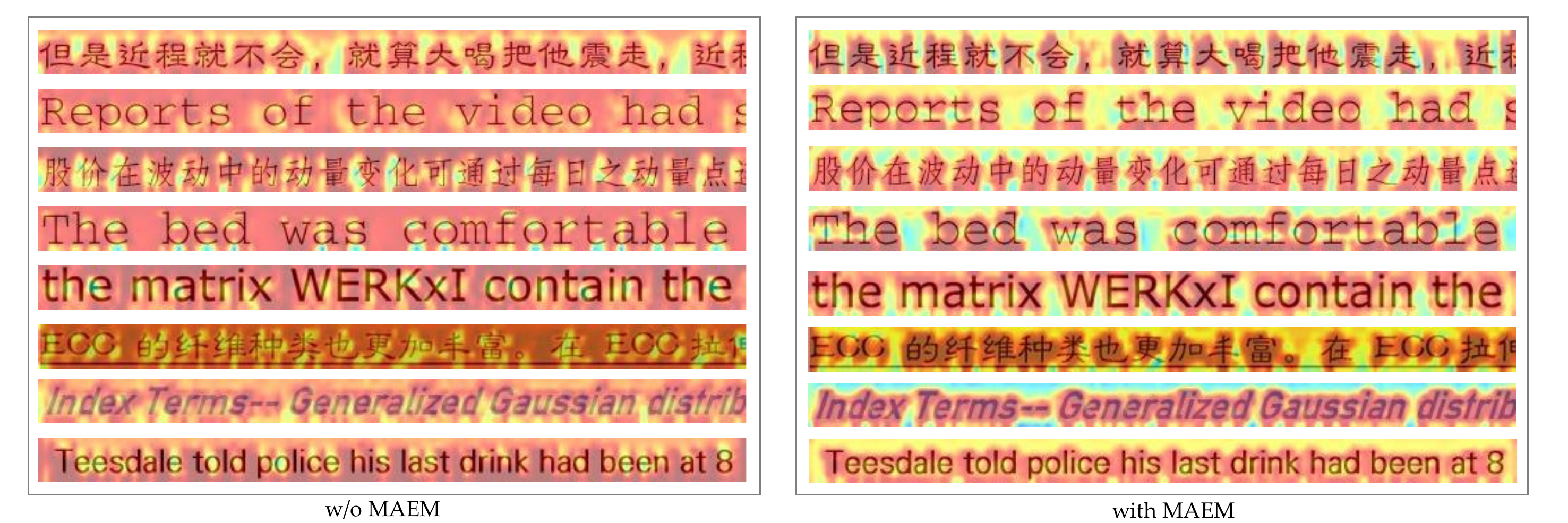}
   \caption{Visualization of attention maps for eight synthetic samples, where darker color indicates higher weights. This is acquired using Grad-CAM~\cite{ref63} and corresponds to the network's first stage output.
   }
   \label{img4}
\end{figure*}

\subsection*{C. Selecting SimSiam as the basic paradigm}
Recently several advanced contrastive frameworks~\cite{ref52,ref53} incorporating negative pairs show superior performance beyond SimSiam. 
Nevertheless, for textual attribute recognition, the inter-samples relation is more subtle and lacks unique background features, while the latter is an essential part of instance discrimination. 
We mainly focus on leveraging pre-training techniques to improve the fine-grained attribute recognition tasks, and experiments fully demonstrate the huge benefits brought by our framework. 
We also visualize the correlation matrices of features (20 classes)  obtained by supervised learning and TaCo (see Fig.2). TaCo enables the model to output representations with smaller inter-class correlations with the supervised version.
With the supervised version, TaCo enables the model to output representations with smaller inter-class correlations.

\subsection*{D. Details about Semantic-Entity Labeling}
Semantic-Entity Labeling aims to identify the types of existing text fields in an image. 
FUNSD~\cite{ref7} is a noisy scanned document dataset consisting of 199 fully annotated forms with 9707 entities, of which the training set includes 149 forms and the test set includes 49 forms. 
The implementation details are consistent with~\cite{ref20} except for the feature fusion operation mentioned in the main text. 
The evaluation metrics are precision and recall. As shown in Table 1, the inclusion of attribute information leads to a 1.49 precision improvement.

\subsection*{E. More Visualization Results}
We show more visualization results based on the TaCo system in Fig.3 and Fig.4. It is observed that TaCo can provide reliable suggestions even in complex scenarios. More, the attention maps shown in Fig.5 imply that the existence of MAEM allows the network to pay more attention to local traits.

{\small
\bibliographystyle{ieee_fullname}
\bibliography{egbib}
}

\end{document}